\documentclass[letterpaper]{article}
\usepackage{ijcai09}

\usepackage{times}

\usepackage{wrapfig,subfigure}
\usepackage{amsmath,amssymb,algorithm,algorithmic}
\usepackage{graphicx} 
\usepackage{epsfig} 
\newcommand{\reals}{\mathbb{R}}
\newcommand{\T}{{}^\top}
\newcommand{\card}[1]{\left\vert #1 \right\vert}
\renewcommand{\vec}{\mathbf}
\newcommand{\pinch}{}

\title{Streamed Learning: One-Pass SVMs}
\author{Piyush Rai, Hal Daum\'e III, Suresh Venkatasubramanian \\
University of Utah, School of Computing\\
\texttt{\{piyush,hal,suresh\}@cs.utah.edu}}

\begin{document}

\maketitle

\begin{abstract}
We present a streaming model for large-scale classification (in the context of $\ell_2$-SVM)  by leveraging connections between learning and computational geometry. The streaming model imposes the constraint that only a single pass over the data is allowed. The $\ell_2$-SVM is known to have an equivalent formulation in terms of the minimum enclosing ball (MEB) problem, and an efficient algorithm based on the idea of \emph{core sets} exists (CVM) \cite{cvm}. CVM learns a $(1+\varepsilon)$-approximate MEB for a set of points and yields an approximate solution to corresponding SVM instance. However CVM works in batch mode requiring multiple passes over the data. This paper presents a single-pass SVM which is based on the minimum enclosing ball of streaming data. We show that the MEB updates for the streaming case can be easily adapted to learn the SVM weight vector in a way similar to using online stochastic gradient updates. Our algorithm performs polylogarithmic computation at each example, and requires very small and constant storage. Experimental results show that, even in such restrictive settings, we can learn efficiently in just one pass and get accuracies comparable to other state-of-the-art SVM solvers (batch and online). We also give an analysis of the algorithm, and discuss some open issues and possible extensions.
\end{abstract}

\section{Introduction}
\label{intro}
Learning in a streaming model poses the restriction that we are constrained both in terms of time, as well as storage. Such scenarios are quite common, for example, in cases such as analyzing network traffic data, when the data arrives in a streamed fashion at a very high rate. Streaming model also applies to cases such as disk-resident large datasets which cannot be stored in memory. Unfortunately, standard learning algorithms do not scale well for such cases. To address such scenarios, we propose applying the \emph{stream model} of computation \cite{datastreamsurvey} to supervised learning problems. In the stream model, we are allowed only one pass (or a small number of passes) over an ordered data set, and polylogarithmic storage and polylogarithmic computation per element.  In spite of the severe limitations imposed by the streaming framework, streaming algorithms have been successfully employed in many different domains \cite{streamcluster}. Many of the problems in geometry can be adapted to the streaming setting and since many learning problems have equivalent geometric formulations, streaming algorithms naturally motivate the development of efficient techniques for solving (or approximating) large-scale batch learning problems.

In this paper, we study the application of the stream model to the problem of maximum-margin classification, in the context of $\ell_2$-SVMs \cite{vapnik98:statlearnth,crist00:introsvm}.  Since the support vector machine is a widely used classification framework, we believe success here will encourage further research into other frameworks.  SVMs are known to have a natural formulation in terms of the minimum enclosing ball problem in a high dimensional space \cite{cvm,bvm}. This latter problem has been extensively studied in the computational geometry literature and admits natural streaming algorithms \cite{chanstream,ahvstream}. We adapt these algorithms to the classification setting, provide some extensions, and outline some open issues.  Our experiments show that we can learn efficiently in just one pass and get competetive classification accuracies on synthetic and real datasets.

\section{Scaling up SVM Training}
\label{sec:supp-vect-mach}
Support Vector Machines (SVM) are maximum-margin kernel-based linear classifiers \cite{crist00:introsvm} that are known to provide provably good generalization bounds \cite{vapnik98:statlearnth}. Traditional SVM training is formulated in terms of a quadratic program (QP) which is typically optimized by a numerical solver. For a training size of $N$ points, the typical time complexity is $O(N^3)$ and storage required is $O(N^2)$ and such requirements make SVMs prohibitively expensive for large scale applications.  Typical approaches to large scale SVMs, such as chunking \cite{vapnik98:statlearnth}, decomposition methods \cite{libsvm} and SMO \cite{smo} work by dividing the original problem into smaller subtasks or by scaling down the training data in some manner \cite{svmhierclust,rsvm}.  However, these approaches are typically heuristic in nature: they may converge very slowly and do not provide rigorous guarantees on training complexity \cite{cvm}.
There has been a recent surge in interest in the online learning literature for SVMs due to the success of various gradient descent approaches such as stochastic gradient based methods \cite{sg} and stochastic sub-gradient based approaches\cite{pegasos}. These methods solve the SVM optimization problem iteratively in steps, are quite efficient, and have very small computational requirements. Another recent online algorithm LASVM \cite{lasvm} combines online learning with active sampling and yields considerably good performance doing single pass (or more passes) over the data. However, although fast and easy to train, for most of the stochastic gradient based approaches, doing a single pass over the data does not suffice and they usually require running for several iterations before converging to a reasonable solution.

\section{Two-Class Soft Margin SVM as the MEB Problem}
\label{sec:two-class-soft}

A minimum enclosing ball (MEB) instance is defined by a set of points $\vec x_1$, ..., $\vec x_N \in \reals^D$ and a metric $d : \reals^D \times \reals^D \rightarrow \reals^{\geq 0}$. The goal is to find a point (the \emph{center}) $\vec c \in \reals^D$ that minimizes the radius $R = \max_n d(\vec x_n, \vec c)$.

The 2-class $\ell_2$-SVM \cite{cvm} is defined by a hypothesis $f(\vec{x}) = \vec{w}^T\varphi(\vec{x})$, and a training set consisting of $N$ points $\{\vec{z}_n= (\vec{x}_n,y_n)\}_{n=1}^N$ with $y_n \in \{-1,1\}$ and $\vec x_n \in \reals^D$. The primal of the two-classs $\ell_2$-SVM (we consider the unbiased case one---the extension is straightforward) can be written as 
\begin{equation} 
\min_{\vec{w},\xi_i} ||\vec{w}||^2 + C\sum_{i=1,m}\xi_i^2
\end{equation}
\begin{equation} s.t. \ \ y_i(\vec{w}'\varphi(\vec{x}_i)) \geq 1 - \xi_i ,\ \ i=1,...,N \end{equation}

The only difference between the $\ell_2$-SVM and the standard SVM is that the penalty term has the form $(C\sum_n{{\xi_n}^2})$ rather than $(C\sum_n{\xi_n})$. 

We assume a kernel $K$ with associated nonlinear feature map $\varphi$.  We further assume that $K$ has the property $K(\vec{x},\vec{x}) = \kappa$, where $\kappa$ is a fixed constant \cite{cvm}. Most standard kernels such as the isotropic, dot product (normalized inputs), and normalized kernels satisfy this criterion.

Suppose we replace the mapping $\varphi(\vec{x}_n)$ on $\vec{x}_n$  by another nonlinear mapping $\tilde \varphi(\vec{z}_n)$ on $\vec{z}_n$ such that (for unbiased case) 
\begin{equation}\tilde \varphi(\vec{z}_n) = \left[ y_n \varphi(\vec x_n) ;  C^{-1/2} \vec e_n \right]\T \end{equation} The mapping is done in a way that that the label information $y_n$ is subsumed in the new feature map $\tilde \varphi$ (essentially, converting a supervised learning problem into an unsupervised one).  The first term in the mapping corresponds to the feature term and the second term accounts for a regularization effect, where $C$ is the misclassification cost.  $\vec{e}_n$ is a vector of dimension $N$, having all entries as zero, except the $n^{\text{th}}$ entry which is equal to one.

It was shown in \cite{cvm} that the MEB instance $(\tilde\varphi(\vec z_1), \tilde\varphi(\vec z_2), \ldots \tilde\varphi(\vec z_N))$, with the metric defined by the induced inner product, is dual to the corresponding $\ell_2$-SVM instance (1). The weight vector $\vec{w}$ of the maximum margin hypothesis can then be obtained from the center $\vec{c}$ of the MEB using the constraints induced by the Lagrangian \cite{bvm}.

\section{Approximate and Streaming MEBs}
\label{sec:appr-stre-mebs}

The minimum enclosing ball problem has been extensively studied in the computational geometry literature. An instance of MEB, with a metric defined by an inner product, can be solved using quadratic programming\cite{boyd04:_convex_optim}. However, this becomes prohibitively expensive as the dimensionality and cardinality of the data increases; for an $N$-point SVM instance in $D$ dimensions, the resulting MEB instance consists of $N$ points in $N+D$ dimensions. 

Thus, attention has turned to efficient approximate solutions for the MEB.  A $\delta$-approximate solution to the MEB ($\delta > 1$) is a point $\vec c$ such that $\max_n d(\vec x_n,\vec c) \le \delta R^*$, where $R^*$ is the radius of the true MEB solution. For example, A $(1+\epsilon)$-approximation for the MEB can be obtained by extracting a very small subset (of size $O(1/\epsilon)$) of the input called a \emph{core-set} \cite{AHV}, and running an exact MEB algorithm on this set  \cite{badoiuminball}. This is the method originally employed in the CVM \cite{cvm}. \cite{maxmargcore} take a more direct approach, constructing an explicit core set for the (approximate) maximum-margin hyperplane, without relying on the MEB formulation. Both these algorithms take linear training time and require very small storage. Note that a $\delta$-approximation for the MEB directly yields a $\delta$-approximation for the regularized cost function associated with the SVM problem.

Unfortunately, the core-set approach cannot be adapted to a streaming setting, since it requires $O(1/\epsilon)$ passes over the training data. Two one-pass streaming algorithms for the MEB problem are known.  The first \cite{ahvstream} finds a $(1+\epsilon)$ approximation using $O((1/\varepsilon)^{\lfloor D/2 \rfloor})$ storage and $O((1/\varepsilon)^{\lfloor D/2 \rfloor}N)$ time.  Unfortunately, the exponential dependence on $D$ makes this algorithm impractical. At the other end of the space-approximation tradeoff, the second algorithm \cite{chanstream} stores only the center and the radius of the current ball, requiring $O(D)$ space. This algorithm yields a 3/2-approximation to the optimal enclosing ball radius.

\subsection{The StreamSVM Algorithm}

We adapt the algorithm of \cite{chanstream} for computing an approximate maximum margin classifier. The algorithm initializes with a single point (and therefore an MEB of radius zero).  When a new point is read in off the stream, the algorithm checks whether or not the current MEB can enclose this point. If so, the point is discarded. If not, the point is used to suitably update the center and radius of the current MEB. All such selected points define a core set of the original point set.

\begin{figure}[ht]
\vskip 0.2in
\begin{center}
\setlength{\epsfxsize}{1.25in}
\centerline{\epsfbox{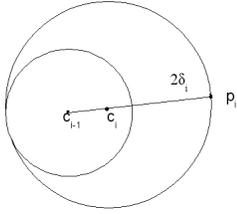}}
\caption{Ball updates}
\label{ballupdate}
\end{center}
\vskip -0.2in
\end{figure}

Let $\vec{p}_i$ be the input point causing an update to the MEB and $\vec{B}_i$ be the resulting ball after the update. From figure \ref{ballupdate}, it is easy to verify that the new center $\vec{c}_i$ lies on the line joining the old center $\vec{c}_{i-1}$ and the new point $\vec{p}_i$. The radius $\vec{r}_i$ and the center $\vec{c}_i$ of the resulting MEB can be defined by simple update equations.
\begin{equation} r_i = r_{i-1} + \delta_i \end{equation}
\begin{equation}||\vec{c}_i - \vec{c}_{i-1}||=\delta_i \end{equation}
Here $2\delta_i = (||\vec{p}_i - \vec{c}_{i-1}|| - r_{i-1})$ is the closest distance of the new point $\vec{p}_i$ from the old ball $\vec{B}_{i-1}$. Using these, we can define a closed-form analytical update equation for the new ball $\vec{B}_i$:
\begin{equation} \vec{c}_i = \vec{c}_{i-1} + \frac{\delta_i}{||\vec{p}_i - \vec{c}_{i-1}||}(\vec{p}_i - \vec{c}_{i-1}) \end{equation}

It can be shown that, for adversarially constructed data, the radius of the MEB computed by the algorithm has a lower-bound of $(1 + \sqrt2)/2$ and a worst-case upper-bound of $3/2$ \cite{chanstream}. 

We adapt these updates in a natural way in the augmented feature space $\tilde\varphi$ (see Algorithm \ref{alg:streamsvm1}). Each selected point belongs to the \emph{core set} for the MEB. The support vectors of the corresponding SVM instance come from this set. It is easy to verify that the update equations for weight vector ($\bf w$) and the margin (R) in StreamSVM correspond to the center and radius updates for the ball in equation 7 and 4 respectively. The $\xi^2$ term is the distance calculation is included to account for the fact that the distance computations are being done in the $D+N$ dimensional augmented feature space $\tilde\varphi$ which, for the linear kernel case, is given by: 
\begin{equation}\tilde \varphi(\vec{z}_n) = \left[ y_n\vec x_n ;  C^{-1/2} \vec e_n \right]\T.\end{equation} 
Also note that, because we perform only a single pass over the data and the $\vec e_n$ components are all mutually orthogonal, we never need to explicitly store them.  The number of updates to the weight vector is limited by the number of core vectors of the MEB, which we have experimentally found to be much smaller as compared to other algorithms (such as Perceptron).  The space complexity of StreamSVM is small since only the weight vector and the radius need be stored.

\subsection{Kernelized StreamSVM}
Although our main exposition and experiments are with linear kernels, it is straightforward to extend the algorithm for nonlinear kernels. In that case, algorithm 1, instead of storing the weight vector \textbf{w}, stores an $N$-dimensional vector of Lagrange coefficients $\mathbf{\alpha}$ initialized as $\left[y_1,\ldots,0\right]$. The distance computation is line 5 are replaced by $d^2 = \sum_{n,m}\alpha_n\alpha_m k(\vec{x}_n,\vec{x}_m) + k(\vec{x}_n,\vec{x}_n) - 2y_n\sum_{m}\alpha_m k(\vec{x}_n,\vec{x}_m) + \xi^2 + 1/C$, and the weight vector updates in line 7 can be replaced by Lagrange coefficients updates $\mathbf{\alpha}_{1:n-1} = \mathbf{\alpha}_{1:n-1}(1 - \frac 1 2 \left(1 - R/d\right))$, $\alpha_n = \frac 1 2 \left(1 - R/d\right)y_n$.

\begin{algorithm}[!ht]
	\caption{StreamSVM}
	\label{alg:streamsvm1}
\begin{algorithmic}[1]
	\STATE {\bfseries Input:} examples $(\vec x_n,y_n)_{n \in 1\dots N}$, slack parameter $C$
	\STATE {\bfseries Output:} weights ($\vec{w}$), radius ($R$), number of support vectors ($M$)
	\STATE Initialize: {$M=1; R=0; \xi^2=1,\vec{w} = y_1\vec{x}_1$}
	\FOR{$n=2$ to $N$}
	\STATE Compute distance to center:\\ {$d = \sqrt{\|\vec{w} - y_n\vec{x}_n\|^2 + \xi^2 + 1/C}$}
	\IF{$d \geq R$}
	\STATE{$\vec{w} = \vec{w} + \frac 1 2 \left(1 - R/d\right)(y_n\vec{x}_n - \vec{w})$}
	\STATE{$R = R+\frac 1 2 (d-R)$}
	\STATE{$\xi^2 = \xi^2 \left[1-\frac 1 2 \left(1 - R/d\right)\right]^2+\left[\frac 1 2 \left(1 - R/d\right)\right]^2$}
	\STATE{$M = M + 1$}
	\ENDIF
	\ENDFOR
\end{algorithmic}
\end{algorithm}

\begin{algorithm}[!ht]
\caption{StreamSVM with lookahead L}\label{alg:streamsvm2}
\begin{algorithmic}[1]
\item[\textbf{Input:}] examples $(\vec x_n,y_n)_{n \in 1\dots N}$, slack parameter $C$, lookahead parameter $L \geq 1$
\item[\textbf{Output:}] weights ($\vec{w}$), radius ($R$), upper bound on number of support vectors ($M$)
\medskip
\STATE Initialize: {$M=1; R=0; \xi^2=1; \vec S = \emptyset; \vec{w} = y_1\vec{x}_1$}
\FOR{$n=2$ to $N$}
\STATE Compute distance to center: \\{$d = \sqrt{\|\vec{w} - y_n\vec{x}_n\|^2 + \xi^2 + 1/C}$}
\IF{$d \geq R$}
\STATE Add example $n$ to the active set: \\{$\vec S = \vec S \cup \{ y_n\vec x_n \}$}
\IF {$\card{\vec{S}} = L$}
\STATE Update $\vec w,R,\xi^2$ to enclose the ball $(\vec w,R,\xi^2)$ and all points in $\vec S$
\STATE{$M = M + L$ ; $\vec{S} = \emptyset$}
\ENDIF
\ENDIF
\ENDFOR
\IF{$\card{\vec S} > 0$}
\STATE Update $\vec w,R,\xi^2$ to enclose the ball $(\vec w,R,\xi^2)$ and all points in $\vec S$
\STATE{$M = M + \card{\vec S}$}
\ENDIF
\end{algorithmic}
\end{algorithm}

\subsection{StreamSVM approximation bounds and extension to multiple balls}
It was shown in \cite{chanstream} that any streaming MEB algorithm that uses only $O(D)$ storage obtains a lower-bound of  $(1 + \sqrt2)/2$ and an upper-bound of 3/2 on the quality of solution (i.e., the radius of final MEB). Clearly, this is a conservative approximation and would affect the obtained margin of the resulting SVM classifier (and hence the classification performance). In order to do better in just a single pass, one possible conjecture could be that the algorithm must \emph{remember} more. To this end, we therefore extended algorithm-\ref{alg:streamsvm1} to simultaneously store $L$ weight vectors (or ``balls''). The space complexity of this algorithm is $L(D+1)$ floats and it still makes only a single pass over the data. In the MEB setting, our algorithm chooses with each arriving datapoint (that is not already enclosed in any of the balls) how the current $L+1$ balls (the $L$ balls plus the new data point) should be merged, resulting again into a set of $L$ balls. At the end, the final set of $L$ balls are merged together to give the final MEB. A special variant of the $L$ balls case is when all but one of the $L$ balls are of zero radius. This amounts to storing a ball of non-zero radius and to keeping a \emph{buffer} of $L$ many data-points (we call this the \emph{lookahead} algorithm - Algorithm \ref{alg:streamsvm2}). Any incoming point, if not already enclosed in the current ball, is stored in the buffer. We solve the MEB problem (using a quadratic program of size $L$) whenever the buffer is full. Note that algorithm \ref{alg:streamsvm1} is a special case of algorithm \ref{alg:streamsvm2} with $L$=1, with the MEB updates available in a closed analytical form (rather than having to solve a QP).

Algorithm \ref{alg:streamsvm1} takes linear time in terms of the input size. Algorithm \ref{alg:streamsvm2} which uses a lookahead of $L$ solves a quadratic program of size $L$ whenever the buffer gets full. This step takes $O(L^3)$ times. The number of such updates is $O(N/L)$ (in practice, it is considerably less than $N/L$) and thus the over all complexity for the lookahead case is $O(NL^2)$. For small lookaheads, this is roughly $O(N)$.

\section{Experiments}
\label{exper}
We evaluate our algorithm on several synthetic and real datasets and compare it against several state-of-the-art SVM solvers. We use 3 crieria for evaluations: a) Single-pass classification accuracies compared against single-pass of online SVM solvers such as iterative sub-gradient solver Pegasos \cite{pegasos}, LASVM \cite{lasvm}, and Perceptron \cite{rosenb:perceptron}. b) Comparison with CVM \cite{cvm} which is a batch SVM algorithm based on the MEB formulation. c) Effect of using lookahead in StreamSVM. For fairness, all the algorithms used a linear kernel.

\subsection{Single-Pass Classification Accuracies}
The single-pass classification accuracies of StreamSVM and other online SVM solvers are shown in table-\ref{tab:res1}. Details of the datasets used are shown in table-\ref{tab:res1}. To get a sense of how good the single-pass approximation of our algorithm is, we also report the classification accuracies of batch-mode (i.e., all data in memory, and multiple passes) libSVM solver with linear kernel on all the datasets. The results suggest that our single-pass algorithm StreamSVM, using a small reasonable lookahead, performs comparably to the batch-mode libSVM, and does significantly better than a single pass of other online SVM solvers.

\begin{table*}[!htbp]
\centering
\begin{tabular}{|l|c|cc|c|c|cc|c|cc|}
\hline
               &  & \multicolumn{2}{c|}{\bf \# Examples} & \multicolumn{1}{c|}{\bf libSVM} & \multicolumn{1}{c|}{\bf Perceptron} & \multicolumn{2}{c|}{\bf Pegasos} & \multicolumn{1}{c|}{\bf LASVM} &\multicolumn{2}{c|}{\bf StreamSVM} \\
{\bf Data Set} & \bf Dim & {\bf Train} & {\bf Test} & {\bf{(batch)}}& & {k = 1} & {k = 20} &  &{\bf {Algo-1}}  & {\bf{Algo-2}} \\
\hline
Synthetic A      & 2 & 20,000 & 200 & 96.5 & 95.5 & 83.8 & 89.9 & 96.5 &  95.5 &  \bf 97.0 \\
Synthetic B      & 3 & 20,000 & 200 & 66.0 & 68.0 & 57.05 & 65.85 & 64.5 & 64.4 &  \bf 68.5 \\
Synthetic C      & 5 & 20,000 & 200 & 93.2 & 77.0 & 55.0 & 73.2 & 68.0 &  73.1 &  \bf 87.5 \\
Waveform      & 21 & 4000 & 1000 & 89.4 & 72.5 & 77.34 & 78.12 & 77.6 &  74.3 &  \bf 78.4 \\
MNIST (0vs1)   & 784 & 12,665 & 2115 &  99.52 & 99.47 & 95.06 & 99.48 & 98.82 & 99.34 &  \bf 99.71\\
MNIST (8vs9)   & 784 & 11,800 & 1983 &  96.57 & \bf 95.9 & 69.41 & 90.62 & 90.32 & 84.75 & 94.7 \\
IJCNN & 22 & 35,000 & 91,701 &  91.64 & 64.82 & 67.35 & 88.9 & 74.27  & 85.32 & \bf 87.81 \\
w3a & 300 & 44,837 & 4912 &  98.29 & 89.27 & 57.36 & 87.28 & \bf 96.95 & 88.56 &  89.06\\
\hline
\end{tabular}
\caption{\small{Single pass classification accuracies of various algorithms (all using linear kernel). The synthetic datasets (A,B,C) were generated using normally distributed clusters, and were of about 85\% separability. libSVM, used as the absolute benchmark, was run in batch mode (all data in memory). StreamSVM Algo-2 used a small lookahead ($\sim$10). Note: We make the Pegasos implementation do a single sweep over data and have a user chosen block size $k$ for subgradient computations (we used k=1, and k=20 akin to using a lookahead of 20). Perceptron and LASVM are also run for a single pass and do not need block sizes to be specified. All results are averaged over 20 runs (w.r.t. random orderings of the stream)}\label{tab:res1}} 
\end{table*}

\subsection{Comparison with CVM}
\begin{figure}[!htbp]
   \centering
\includegraphics[width=3in]{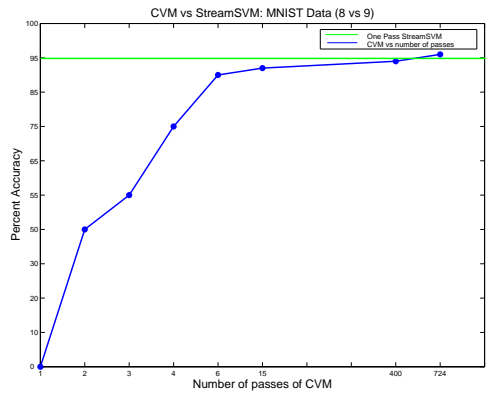}

\caption{\small{MNIST 8vs9 data: Number of passes CVM takes before achieving comparable single-pass accuracy of StreamSVM. X axis represents number of passes of CVM and Y axis represents the classification accuracy.}}
\label{fig:cvm}     
\end{figure}

We compared our algorithm with CVM which, like our algorithm, is based on a MEB formulation. CVM is highly efficient for large datasets but it operates in batch mode, making one pass through the data for each core vector.  We are interested in knowing how many passes the CVM must make over the data before it achieves an accuracy comparable to our streaming algorithm. For that purpose, we compared the accuracy of our single-pass StreamSVM against two and more passes of CVM to see how long does it take for CVM to beat StreamSVM (we note here that CVM requires at least two passes over the data to return a solution).  We used a linear kernel for both. Shown in Figure~\ref{fig:cvm} are the results on MNIST 8vs9 data and it turns out that it takes several hundreds of passes of CVM to beat the single pass accuracy of StreamSVM. Similar results were obtained for other datasets but we do not report them here due to space limitations. 

\begin{figure}[!htbp]
   \centering
\includegraphics[width=3in]{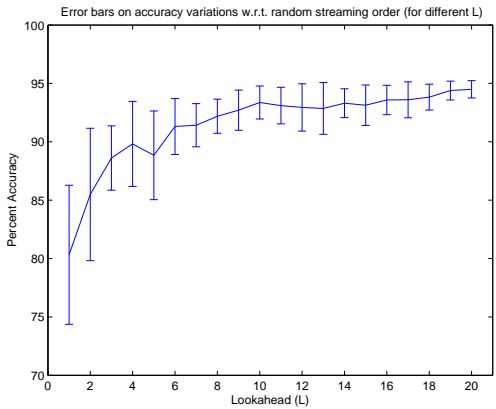}
\caption{\small{Single-pass with varying lookahead on MNIST 8vs9 data: Performance w.r.t random ordering of streaming. X axis represents the lookahead parameter and Y axis represents classification accuracy. Verticle bars represent the standard deviations in accuracies for a given lookahead.}}
\label{fig:varyL}
\end{figure} 

\subsection{Effect of Lookahead}
We also investigated the effect of doing higher-order lookaheads on the data. For this, we varied $L$ (the lookahead parameter) and, for each $L$, tested Algorithm~\ref{alg:streamsvm2} on 100 random permutations of the data stream order, also recording the standard deviation of the classification accuracies with respect to the data-order permutations. Note that the algorithm still performs a single pass over the data.  Figure~\ref{fig:varyL} shows the results on the MNIST 8vs9 data (similar results were obtained for other datasets but not shown due to space limitations).  In this figure, we see two effects. Firstly, as the lookahead increase, performance goes up.  This is to be expected since in the limit, as the lookahead approaches the data set size, we will solve the exact MEB problem (albeit at a high computational cost). The important thing to note here is that even with a small lookahead of $10$, the performance converges.  Secondly, we see that the standard deviation of the result decreases as the lookahead increases.  This shows experimentally that higher lookaheads make the algorithm less susceptible to badly ordered data. This is interesting from an empirical perspective, given that we can show that in theory, any value of $L < N$ cannot improve upon the 3/2-approximation guaranteed for $L=1$.

\section{Analysis, Open Problems, and Extensions}
\label{openissues}
There are several open problems that this work brings up:
\begin{enumerate}
 \item Are the $(1+\sqrt{2})/2$ lower-bound and the $3/2$ upper-bound on MEB radius indeed the best achievable in a single pass over the data?
 \item Is it possible to use a richer geometric structure instead of a ball and come up with streaming variants with provably good approximation bounds?
\end{enumerate}

We discuss these in some more detail here. 

\subsection{Improving the Theoretical Bounds}

One might conjecture that storing more information (i.e., more points) would give better approximation guarantees in the streaming setting. Although the empirical results showed that such approaches do result in better classification accuracies, this is not theoretically true in many cases.

For instance, in the adversarial stream setting, one can show that \emph{neither} the lookahead algorithm \emph{nor} its more general case (the multiple balls algorithm) improves the bounds given by the simple no-lookahead case (Algorithm-\ref{alg:streamsvm1}).  In particular, one can prove an identical upper- and lower-bound for the lookahead algorithm as for the no-lookahead algorithm.  To obtain the $3/2$-upper bound result, one can show a nearly identical construction as to \cite{chanstream} where $L-1$ points are packed in a small, carefully constructed cloud the boundary of the true MEB.

\begin{figure}[ht]
\centering
\includegraphics{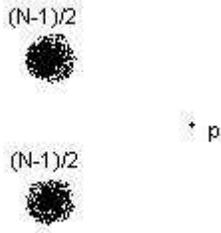}
\caption{\small{An adversarially constructed setting.}}
\label{fig:advstream}
\end{figure}

Alternatively, one can analyze these algorithms in the random stream setting. Here, the input points are chosen adversarially, but their \emph{order} is permuted randomly.  The lookahead model is not strengthened in this setting either: we can show both that the lower bound for no-lookahead algorithms, as well as the 3/2-upper bound for the specific no-lookahead algorithm described, generalize. For the former, see Figure~\ref{fig:advstream}. We place $(N-1)/2$ points around $(0,1)$ and $(N-1)/2$ points around $(0,-1)$ and one point at $(1+\sqrt{2},0)$.  The algorithm will only beat the $(1+\sqrt{2})/2$ lower bound if the singleton appears in the first $L$ points, where $L$ is the lookahead used.  Assuming the lookahead is polylogarithmic in $N$ (which must be true for a streaming algorithm), this means that as $N \longrightarrow \infty$, the probability of a better bound tends toward zero.  Note, however, that this applies only to the lookahead model, not to the more general multiple balls model, where it \emph{may} be possible to obtain a tighter bounds in the random stream setting.

\subsection{Ellipsoidal Balls}

Instead of using a minimum enclosing ball of points, an alternative could be to use a minimum volume ellipsoid (MVE) \cite{mve05}. An ellipsoid in $\mathbb{R}^D$ is defined as follows: $\{ \vec{x} : (\vec{x}-\vec{c})'\vec{A}(\vec{x}-\vec{c}) <= 1\}$ where $\vec{c} \in \mathbb{R}^D$, $\vec{A} \in \mathbb{R}^{DxD}$, and $\vec{A} \succeq 0$ (positive semi-definite). 

Note that a ball, upon inclusion of a new point, expands equally in all dimensions which may be unnecessary. On the other hand, an ellipsoid can have several axes and scales of variations (modulated by the covariance matrix $\vec{A}$). This allows the ellipsoid to expand only along those directions where needed. In addition, such an approach can also be seen along the lines of confidence weighted linear classifiers \cite{confweighted}. The confidence weighted (CW) method assumes a Gaussian distribution over the space of weight vectors and updates the  mean and covariance parameters upon witnessing each incoming example. Just as CW maintains the models uncertainty using a Gaussian, an ellipsoid generaization can model the uncertainty using the covariance matrix $\vec{A}$. Recent work has shown that there exist streaming possibilities for MVE \cite{streammve08}. The approximation gaurantees, however, are very conservative. It would be interesting to come up with improved streaming algorithms for the MVE case and adapt them for classification settings.

\pinch
\section{Conclusion}
\label{conclude}
\pinch Within the streaming framework for learning, we have presented an efficient, single-pass $\ell_2$-SVM learning algorithm using a streaming algorithm for the minimum enclosing ball problem. We have also extended this algorithm to use a \emph{lookahead} to increase robustness against poorly ordered data.  Our algorithm, StreamSVM, satisfies a proven theoretical bound: it provides a $\left(\frac 3 2\right)$-approximation to the optimal solution.  Despite this conservative bound, our algorithm is experimentally competitive with alternative techniques in terms of accuracy, and learns much simpler solutions. We believe that a careful study of stream-based learning would lead to high quality scalable solutions for other classification problems, possibly with alternative losses and with tighter approximation bounds.

\appendix
\small
\bibliographystyle{named}
\bibliography{ijcai09}

\end{document}